\documentclass{article}

\usepackage{arxiv}

\usepackage{forest}
\useforestlibrary{edges}
\usepackage{subcaption}
\usepackage{threeparttable}
\usepackage{amsmath}
\usepackage{lipsum}
\usepackage{multirow}
\usepackage{blindtext}
\usepackage{amssymb}
\usepackage{soul}
\usepackage[inkscapelatex=false]{svg}
\usepackage{url}
\usepackage{geometry, color, graphicx, float}
\usepackage[colorlinks]{hyperref}
\usepackage{acronym}
\usepackage{natbib}
\usepackage{doi}

\title{FungalZSL: Zero-Shot Fungal Classification with Image Captioning Using a Synthetic Data Approach}

\author{ \href{https://orcid.org/0000-0000-0000-0000}{\includegraphics[scale=0.06]{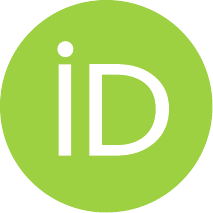}\hspace{1mm}Anju Rani}\thanks{Use footnote for providing further
		information about author (webpage, alternative
		address)---\emph{not} for acknowledging funding agencies.} \\
	Department of Energy Technology\\
	Aalborg University\\
	Esbjerg, Denmark 6700 \\
	\texttt{aran@energy.aau.dk} \\
	\And
	\href{https://orcid.org/0000-0000-0000-0000}{\includegraphics[scale=0.06]{orcid.pdf}\hspace{1mm}Daniel O.~Arroyo} \\
	Department of Energy Technology\\
	Aalborg University\\
	Esbjerg, Denmark 6700 \\
	\texttt{doa@energy.aau.dk} \\
	\And
	\href{https://orcid.org/0000-0000-0000-0000}{\includegraphics[scale=0.06]{orcid.pdf}\hspace{1mm}Petar Durdevic} \\
	Department of Energy Technology\\
	Aalborg University\\
	Esbjerg, Denmark 6700 \\
	\texttt{pdl@energy.aau.dk} \\
}

\renewcommand{\shorttitle}{FungalZSL: Zero-Shot Fungal Classification with Image Captioning Using a Synthetic Data}

\hypersetup{
pdftitle={A template for the arxiv style},
pdfsubject={q-bio.NC, q-bio.QM},
pdfauthor={David S.~Hippocampus, Elias D.~Striatum},
pdfkeywords={First keyword, Second keyword, More},
}

\begin{document}
\maketitle

\begin{abstract}
The effectiveness of zero-shot classification in large vision-language models (VLMs), such as Contrastive Language-Image Pre-training (CLIP), depends on access to extensive, well-aligned text-image datasets. In this work, we introduce two complementary data sources: one generated by large language models (LLMs) to describe the stages of fungal growth and another comprising a diverse set of synthetic fungi images. These datasets are designed to enhance CLIP’s zero-shot classification capabilities for fungi-related tasks. To ensure effective alignment between text and image data, we project them into CLIP’s shared representation space, focusing on different fungal growth stages. We generate text using LLaMA3.2 to bridge modality gaps and synthetically create fungi images. Furthermore, we investigate knowledge transfer by comparing text outputs from different LLM techniques to refine classification across growth stages.

An up-to-date repository accompanies this paper, including the dataset and resources discussed. \href{https://data.mendeley.com/datasets/rw6ndgyrd7/1}{Synthetic Fungi Generation}
\end{abstract}

% keywords can be removed

\section{Introduction}
The emergence of large vision-language models (VLMs), such as CLIP (Contrastive Language-Image Pretraining) \citep{radford2021learning}, BLIP (Bootstrapping Language-Image Pretraining) \citep{li2022blip}, and ALIGN (A Large-scale Image and Noisy-text embedding) \citep{jia2021scaling}, represents a pivotal advancement in representation learning. These models leverage extensive datasets of image-text pairs to create shared embeddings that bridge the gap between visual and natural language domains \citep{min2023recent, minaee2024large, mehta2024catlip}. This bridging facilitates various downstream tasks, including zero-shot classification, where text descriptions are effectively matched to images. Despite their strengths, these models often encounter limitations in encoding detailed visual attributes within fine-grained domains, typically distinguishing categories at a high level. Additionally, the collection of large, domain-specific image-text datasets, particularly for specialized areas like fungi, presents significant resource challenges that complicate the training of these models.

\begin{figure}[H]
\centering
\includegraphics[width=\linewidth]{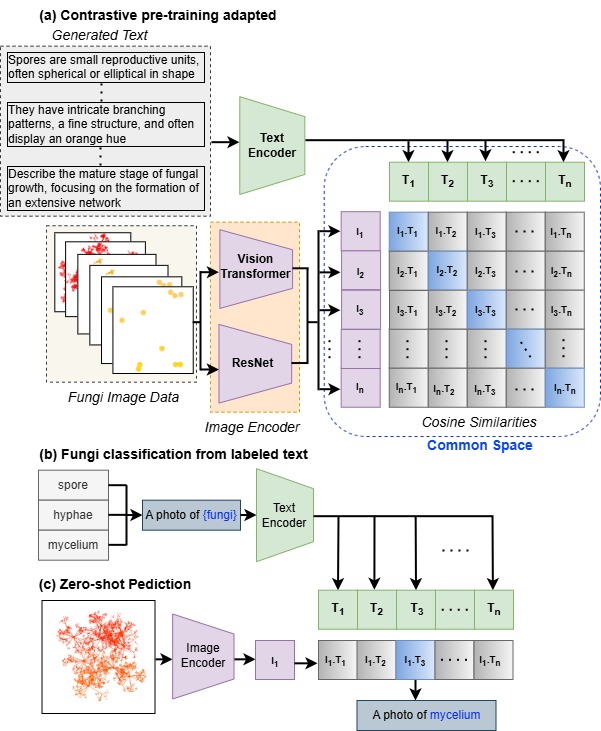}
\caption{Overview of the adapted CLIP model.}
\label{Fig1}
\end{figure}

\textbf{VLMs} such as CLIP \citep{radford2021learning} are designed to learn joint representations of images and text by training on large datasets of image-text pairs. By utilizing separate encoders for text and images, CLIP maps each modality into a shared embedding space where related pairs are positioned closely, while unrelated pairs are separated. This alignment is achieved using a contrastive loss (CL) function, which maximizes the similarity between correct image-text pairs and minimizes it for mismatched pairs within each training batch. This framework enables CLIP to effectively manage large-scale, weakly supervised data and facilitates zero-shot learning, allowing it to classify images into unseen categories by matching them to textual prompts (e.g., “a photo of a [name of class]”). Consequently, CLIP is highly adaptable for transfer learning, with pre-trained encoders that can quickly adjust to new tasks with minimal additional data. The BLIP model \citep{li2022blip} enhances vision-language pretraining through a bootstrapping approach that generates high-quality image-text pairs. It consists of two key components: Bootstrapped Data Generation, which generates relevant captions for images to iteratively refine the dataset for improved semantic alignment, and a contrastive learning (CL) objective that aligns embeddings of related image-text pairs while distancing unrelated ones, paralleling the methods used in CLIP and ALIGN. BLIP has demonstrated strong performance in zero-shot image classification, visual question answering, and image-text retrieval, achieving high accuracy by filtering training data effectively and minimizing dependence on noisy web sources. The ALIGN model \citep{jia2021scaling} capitalizes on vast noisy image-text pairs derived from alt-text annotations on the web. By employing a CL framework, ALIGN achieves robust zero-shot and transfer learning capabilities across diverse tasks without requiring extensive fine-tuning (FT). Its approach effectively navigates the noise inherent in web-based data, leveraging its sheer volume to enhance learning outcomes.

\begin{figure}[ht]
\centering
\includegraphics[width=\linewidth]{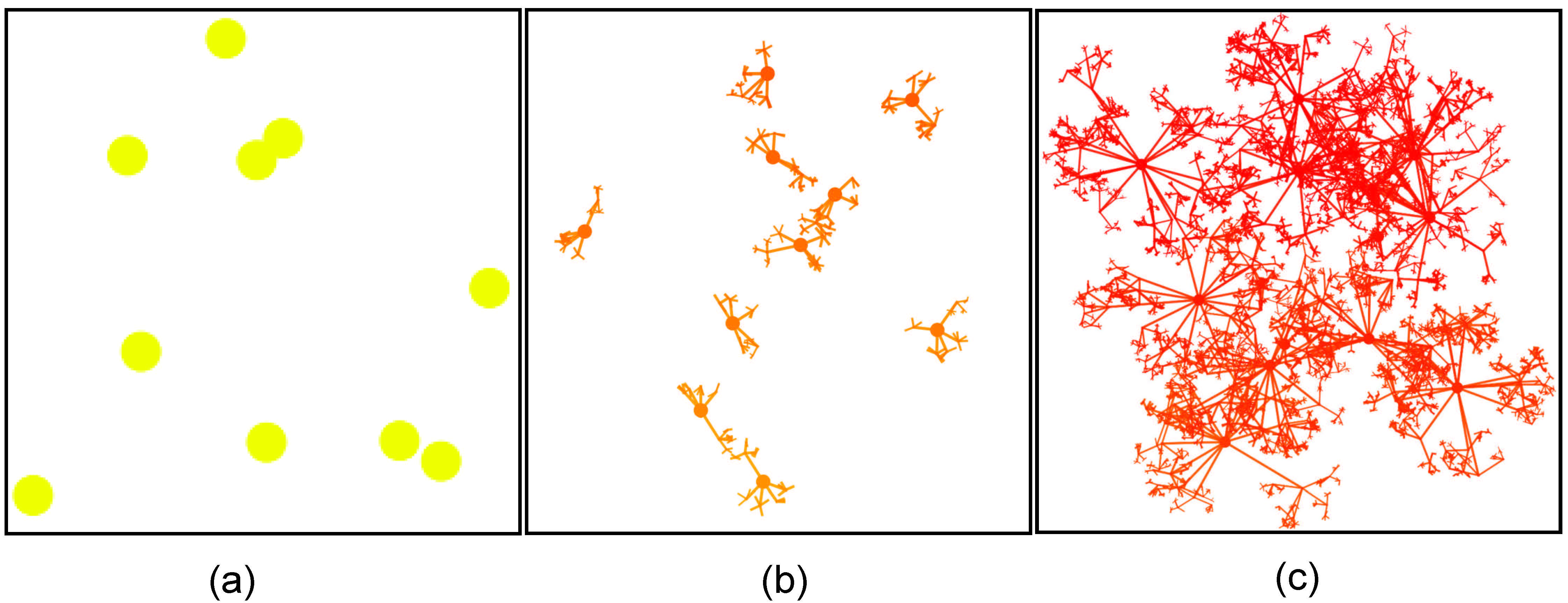}
\caption{Fungi Dataset: (a) Spore, (b) Hyphae, and (c) Mycelium.}
\label{Fig2}
\end{figure}

\begin{figure}[ht]
\centering
\includegraphics[width=\linewidth]{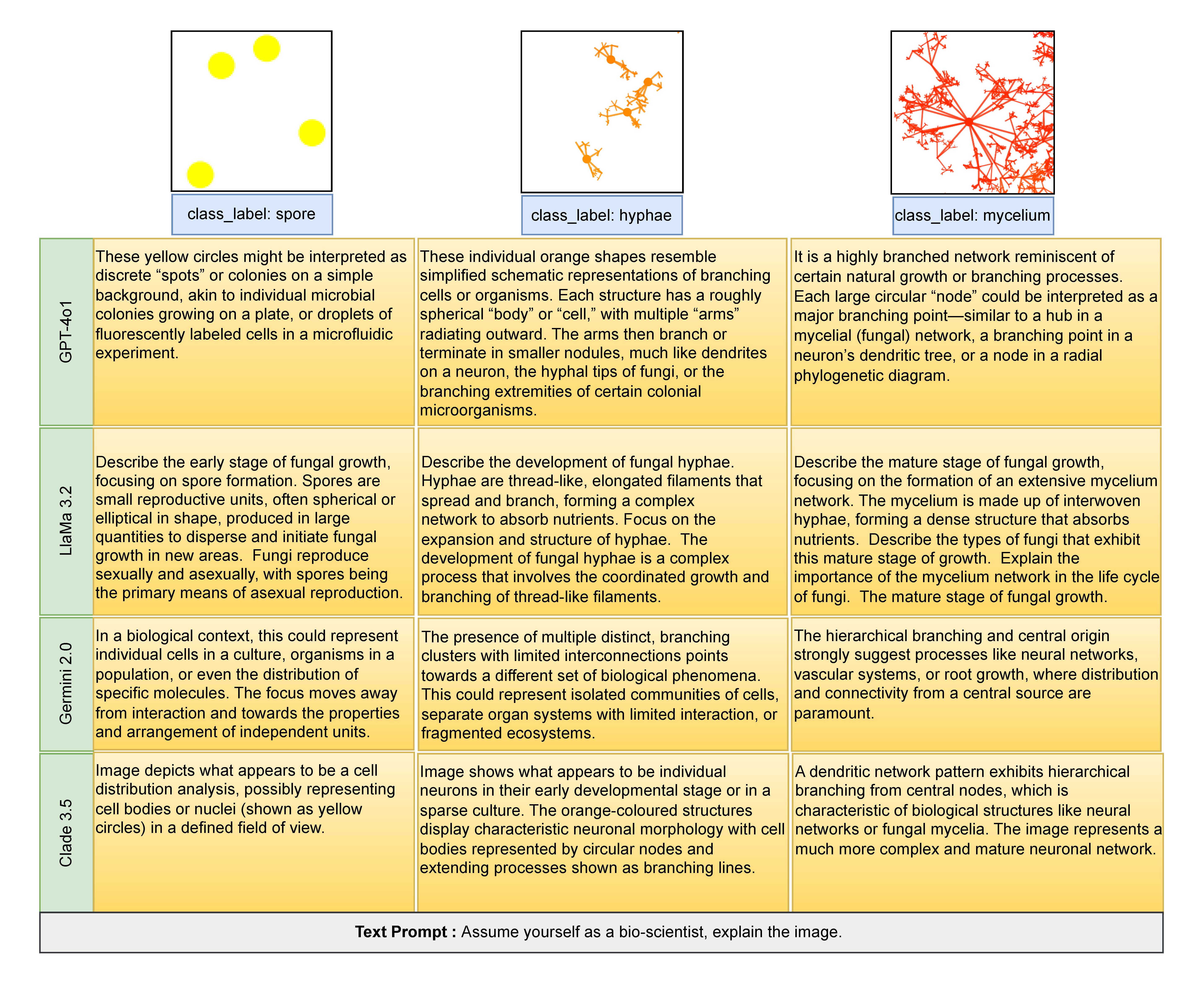}
\caption{Textual description of using various LLMs.}
\label{Fig3}
\end{figure}

Moreover, the classification with Hierarchical Label Sets (CHiLS) framework \citep{novack2023chils} significantly enhances zero-shot image classification performance for models like CLIP by incorporating a hierarchical structure into class labels. This methodology improves model accuracy by substituting coarse class labels with fine-grained sub-classes that better reflect semantic relationships, resulting in notable performance gains—up to 30 \% higher accuracy in datasets with inherent hierarchical structures. CHiLS maintains consistent improvements even in the absence of explicit hierarchical data by generating approximate subclasses through prompts to GPT-3. This adaptable approach requires no additional training costs and can be seamlessly integrated into existing zero-shot classification pipelines. Recently, the CatLIP model \citep{mehta2024catlip} has introduced an innovative shift in vision model pre-training by framing the task on image-text data as a classification problem rather than utilizing CLIP's CL approach. This transformation eliminates the need for pairwise similarity calculations, resulting in a 2.7x acceleration in training on large-scale datasets compared to CLIP. Furthermore, CatLIP enhances data-efficient transfer learning by employing its classification layer to initialize classifiers for target tasks. Unlike CLIP, CatLIP exhibits improved performance with extended training on smaller datasets, successfully avoiding early saturation in performance.

The evolution of \textbf{large language models (LLMs)} has been marked by significant advancements, starting with GPT \citep{radford2018improving}, which introduced the transformer architecture for unsupervised learning on large text data. GPT-2 \citep{radford2019language} scaled up the model to 1.5 billion parameters, showcasing its ability to generate coherent text and perform various tasks without FT, while raising ethical concerns about potential misuse. Building on this, GPT-3 \citep{radford2021learning} further increased the parameter count to 175 billion, demonstrating remarkable capabilities in zero-shot and few-shot learning. The subsequent GPT-4 \citep{openai2023gpt} improved reasoning and contextual understanding, solidifying its reliability for diverse applications. The Mistral model \citep{jiang2023mistral}, designed specifically for efficiency and performance, contributes to the field by delivering high-quality outputs while focusing on reducing computational costs, making it an attractive option for resource-constrained environments. Additionally, models like Gemini \citep{team2023gemini} and Claude \citep{bai2022constitutional} have pushed the boundaries of multimodal reasoning and safety, enhancing AI alignment and interoperability. Meanwhile, LLaMA \citep{touvron2023llama} introduced a family of models optimized for research, ranging from 7 billion to 70 billion parameters, aiming to democratize access to large models while maintaining competitive performance. Building upon these advancements, OpenAI introduced GPT-4o1, a model that emphasizes thinking" before responding, thereby enhancing its problem-solving capabilities in complex domains such as science, coding, and mathematics. Together, these models illustrate the rapid progress in natural language processing (NLP) and their expanding utility across various domains. While much of the prior research has concentrated on refining text descriptions for a wide array of generic classes, the ongoing exploration of LLMs-specific applications and adaptations continues to expand their utility. This focus not only enhances the performance of these models in general tasks but also paves the way for tailored approaches that address more specialized domains and complex user needs. 
To address the limitations of existing models in recognizing fine-grained categories, particularly in the context of fungi growth stages, we propose a novel VLM framework. This work leverages two primary data sources—LLMs and extensive synthetic fungi image datasets—to enhance zero-shot classification performance across these fine-grained domains. Specifically, we generate a synthetic fungi dataset and align it with text descriptions produced by LLMs for each class label, capturing various stages of fungi growth. This approach is effective because the fungi image dataset is fine-grained, unlike broader categories with high-class variation. Additionally, the synthetic fungi dataset uses LLMs to generate scalable descriptions of essential features, such as shape, color, branching, and growth stage. The use of fine-grained labels aids in bridging the gap between closely related images of fungi growth stages, which is crucial for training large VLMs. Figure~\ref{Fig1} illustrates the adapted approach used in this study.

The contributions of this paper are summarized as follows: 
\begin{enumerate} 
\item A novel synthetic image dataset was generated to represent different fungal growth stages capturing fine-grained structural variations.
\item Various LLMs (LLaMA 3.2, GPT-4o1, Claude 3.5, and Gemini 2.0) were used to generate and compare textual descriptions of fungal growth stages. 
\item Systematic FT to improve CLIP performance by bridging the modality gap between synthetic images and text descriptions.
\item Comprehensive evaluation using quantitative metrics, and visual inspections.
\end{enumerate}

The paper is structured as follows: Section 2 outlines the methodology used in this work, detailing the process of generating a synthetic fungi dataset and its corresponding textual descriptions. Section 3 presents the implementation details and the results obtained from our experiments. Finally, Section 4 offers concluding remarks, summarizing our findings and discussing their implications.

\section{Methodology}
We pre-train LLMs on paired multi-modal data to enable multi-modal understanding, specifically focusing on image and text narrations. This process involves training an encoder that projects input data into the text token embedding space of a designated LLM, effectively transforming the LLM's text token embedding space into a joint token embedding space for enhanced interaction between modalities.

The proposed method is evaluated through zero-shot classification on novel fungi classes by assessing the ability of CLIP to recognize fungal growth stages based on textual descriptions generated by different LLMs. To enhance the effectiveness of large VLMs, we generate detailed class descriptions for spore, hyphae, and mycelium using LLaMA 3.2 and test their impact on classification when replaced with descriptions from other LLMs, including GPT-4o1, Claude 3.5, and Gemini 2.0. We employ diverse prompting techniques to improve text-image alignment and assess how well the model generalizes across different textual representations. A summary of the consistent improvements in text generation across LLMs is presented in Figure~\ref{Fig2}.

\subsection{Generation of Fungi Dataset}
Consider a dataset with three classes Class 1: spore (early-stage), Class 2: hyphae (mid-stage), and Class 3: mycelium (mature-stage). Each class has its unique structure and is represented by different characteristics (number of spores, branch structures, given by:

\begin{equation}
C=({C_{s}, C_{h}, C_{m}})
\end{equation}

where \(C_{s}\), \(C_{h}\), and \(C_{m}\) are the three classes of fungi dataset spore, hyphae and mycelium respectively. For each class \(C_{k}\), where \(k\in\) ({spore, hyphae, mycelium}), we define the parameters: \(N_k\): number of structures, \(D_k\): depth of branching, \(L_k\): length of branches, and \(W_k\): width of branches. These parameters control the structure’s appearance in each stage. For each class \(C_{k}\), define a function \(G_k(x,y;\theta_k)\) that generates the structures at random positions \((x,y)\) with the parameter vector \(\theta_k = (N_k, D_k, L_k, W_k)\). 
\begin{enumerate} 
\item \emph{Spore Class Generation}: Define \(G_{s}(x,y; N_{s})\) as a set of circular points \(F\) centered at \((x,y)\) with radius \(r\) representing the spores:
\begin{equation}
G_{s}(x, y; N_{s})= \sum_{i=1}^{N_{s}}F((x_i,y_i),r)
\end{equation}

\item \emph{Hyphae Class Generation}: Define \(G_{h}(x,y; \theta_{h})\) as a central spore with branching lines extending outward, where \(\theta_{h}\) = (\(N_{h}\), \(D_{h}\), \(L_{h}\), \(W_{h}\)). For each spore centre \((x_i,y_i)\), draw branches with recursive depth \(D_{h}\) and initial branch length \(L_{h}\):

\begin{equation}
G_{h}(x, y; \theta_{h})= \sum_{i=1}^{N_{h}}[F((x_i,y_i),r)+ \sum_{d=1}^{D_{h}}B(L_{h}\cdot\alpha^d, W_{h}.\beta^d) ]
\end{equation}

\item \emph{Mycelium Class Generation}: Define \(G_{m}(x,y; \theta_{m}\)) with the most complex branching pattern, where \(\theta_{m}\) = (\(N_{m}\), \(D_{m}\), \(L_{m}\), \(W_{m}\)).

\begin{equation}
G_{m}(x, y; \theta_{m})= \sum_{i=1}^{N_{m}}[F((x_i,y_i),r)+ \sum_{d=1}^{D_{m}}B(L_{m}\cdot\alpha^d, W_{m}\cdot\beta^d) ]
\end{equation}
\end{enumerate}

Finally, to generate the dataset \((D_{final})\), we iterate over all the classes \(C_{k}\) and generate images for each class: 
\begin{equation}
D_{final} = \sum_{k\in({s,h,m})} \sum_{i=1}^{N_{k}}[F((x_i,y_i),r) + \sum_{d=1}^{D_{k}}B(L_{k}\cdot\alpha^d, W_{k}\cdot\beta^d)]
\end{equation}

Consider a dataset \(D_{final} = \{(q_i, r_i) \mid i=1,\dots,N\}\), where \(q_i\in X\) represents images and \(r_i\in Y\) represents corresponding text descriptions. Large VLMs like CLIP use an image encoder \((I)\) and a text encoder \((T)\) such that \(I(q)\approx T(r)\) when an image \(q\) matches a text description \(r\). In this work, we fine-tune CLIP using a synthetically generated fungi dataset, where the text descriptions are produced by LLMs. The dataset is divided into training, validation, and test sets, encompassing three distinct class labels—\textit{spore}, \textit{hyphae}, and \textit{mycelium}—which correspond to different stages of the fungal life cycle. The generated synthetic fungi dataset is open-source and available at \citep{FungalZSL}. 

The generated fungi dataset is designed to realistically represent each growth stage by incorporating subtle overlaps between stages and unique color gradients for each type. To capture the natural progression of growth, each stage includes traces of the previous one  \citep{rani2025synthetic}. For instance, hyphae images contain a few spores, suggesting that not all cells have fully transitioned to hyphae. Similarly, mycelium images reflect the incomplete transition to mature mycelium by including some hyphae. The specific color gradients used for each fungi stage are as follows:

\begin{enumerate} 
\item Spore: Features a gradient from bright yellow to orange, symbolizing the earliest growth phase. 
\item Hyphae: Progresses from orange to a deeper orange, indicating further development. 
\item Mycelium: Shifts from red-orange to deep red, representing a mature growth phase. 
\end{enumerate}

The image generation is structured over a 100 s timeline, with each stage represented by a specific temperature range. 
\begin{enumerate} 
\item Spore Generation: This phase maps temperatures from 300 K to 330 K, covering the initial formation of spores. 
\item Hyphae Growth: The temperature increases from 330 K to 370 K, aligning with the development of hyphae. 
\item Mycelium Formation: Temperatures rise from 370 K to 400 K, culminating in the mature mycelium stage. 
\end{enumerate}

This structured approach provides realistic progression across stages, with carefully selected colors and temperature mappings to enhance the visual and temporal differentiation of each growth phase \citep{FungalZSL}. Figure~\ref{Fig2} illustrates an example of the generated fungi dataset.

The proposed model is evaluated on unseen fungi classes by calculating the similarity score, where for any image \(q\) with a textual description \(r\), the similarity score \(S(q,r)\) is computed as the cosine similarity between the image encoder's output \(I(q)\) and the text encoder's output \(T(r)\).

\subsection{Generation of Textual Description}
The textual descriptions for each class are generated based on prompts tailored to the specific fungal growth stage, emphasizing attributes such as color, structure, development, and function. Here, \([class]\) represents the class label for each of the \(k\) classes in the fungi dataset: spore, hyphae, and mycelium. The LLM generates \(r_k\) textual descriptions for each class \(k\), ensuring diversity and relevance through probabilistic sampling techniques. Prompts such as “Describe the fungal growth stage \([class]\), focusing on its key biological characteristics,” guide the model to produce detailed descriptions. These descriptions take the form of contextual explanations, such as “\([class]\) characterized by \([characteristics]\),” resulting in a set of descriptions \(r^k\) for each class \(k\). The impact of these characteristic-based descriptions has led to notable improvements, as shown in Figure~\ref{Fig3}. 

The class-level textual descriptions used in training were generated utilizing LlaMA 3.2 and then validated with various other APIs:(a) GPT-4o1, (b) Gemini 2.0, and (c) Claude 3.5. We set the temperature parameter to 0.9, allowing the models to incorporate more randomness and produce diverse textual descriptions for each image class across the dataset. For each class \(C\), descriptions are generated using predefined prompts. Each caption satisfies constraints on length \(L\), with \(L_{min}\leqslant L\leqslant L_{max}\). For \(N\) descriptions with batch size \(B\), the captions are generated over \([N/B]\) batches, given by:
\begin{equation}
r(C,N) = \bigcup_{k=1}^{[N/B]}r_{k}(C,B)
\end{equation}
where \(r_{k}(C, B)\) is the set of captions in the \(k\)-th batch.

\begin{figure}[H]
\centering
\includegraphics[width=\linewidth]{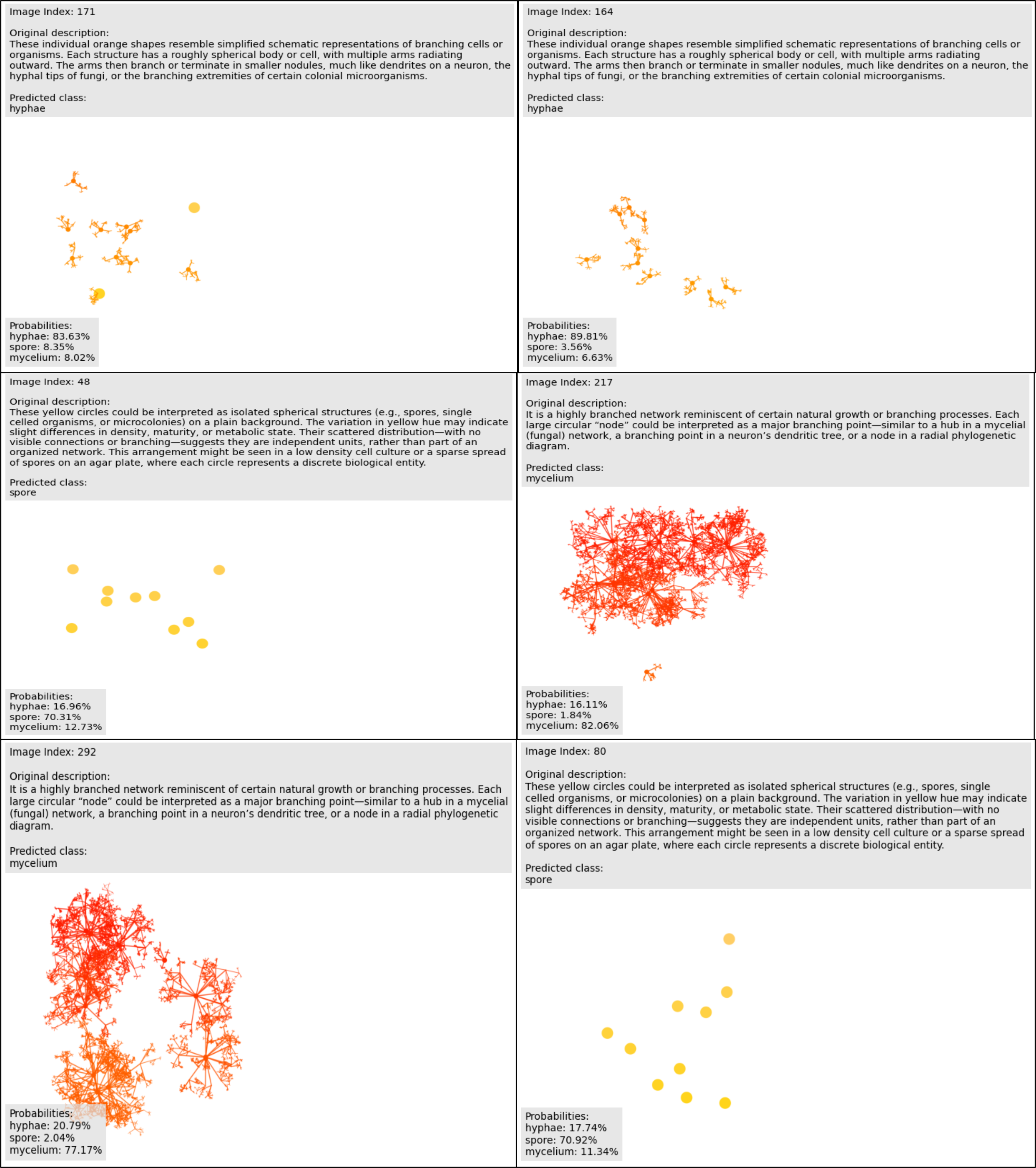}
\caption{Zero-shot classification with proposed CLIP model: (a) top: GPT-4o1, (b) centre: Claude 3.5, and (c) bottom: Gemini 2.0}
\label{Fig4}
\end{figure}

\subsection{CLIP Fine-tuning}
Fine-tuning encompasses a range of techniques aimed at adjusting the parameters of a pre-trained model to optimize its performance on specific downstream tasks. As mentioned in Section 2, for \(q^k\) set of images we have \(r^k\) set of texts for  \(k \in D_{final}\) in the fungi training set. In this work, fine-tuning is performed by continuing CL with image-text pairs. We initialize CLIP’s model (RN50, ViT-B/32, ViT-B/16, and ViT-L/14@336px) and optimize the contrastive loss. The loss function for image and text is given by:
\begin{equation}
L_{image} = -\frac{1}{N}\sum_{q=1}^{N}\frac{1}{\left| P_{q}\right|}\sum_{r\in P_{q}}^{}log\frac{e^{S_{q,r/\tau}}}{\sum_{r=1}^{N}e^{S_{q,r/\tau}}}
\end{equation}

and 

\begin{equation}
L_{text} = -\frac{1}{N}\sum_{r=1}^{N}\frac{1}{\left| P_{r}\right|}\sum_{q\in P_{r}}^{}log\frac{e^{S_{q,r/\tau}}}{\sum_{r=1}^{N}e^{S_{q,r/\tau}}}
\end{equation}

where \(\tau\) represent learnable temperature parameter, \(S_{q,r}\) represents the correct similarity metric (e.g., cosine similarity) between image \emph{q} and text \emph{r}, \(P_q\) represents the set of positive text matches for image \emph{q} while and \(P_r\) represents the set of positive image matches for text \emph{r}. The final contrastive loss (CL) is given by:
\begin{equation}
L_{total} = L_{image}+ L_{text}
\end{equation}

\begin{table}[h]
\caption{Recall@1 qualitative results.}
\centering
\begin{tabular}{p{0.28\linewidth}p{0.22\linewidth}p{0.21\linewidth}p{0.1\linewidth}}
\hline
\textbf{CLIP Model} & \textbf{Learning rate} & \textbf{Iteration rate} & \textbf{Value}\\
\hline
RN50             & 1e-6    & 2.32 s/it     & 0.333          \\
ViT-B/16         & 1e-6    & 15.95 s/it    & 0.913          \\
ViT-B/32         & 1e-6    & 15.50 s/it    & 0.670          \\
ViT-L/14@336px   & 1e-6    & 3.86 s/it     & \textbf{0.973} \\
\hline
\end{tabular}
\label{tbl1}
\end{table}

\section{Results and Discussion}
This section assesses the proposed method against baseline models under various settings. The evaluation focuses on the synthetic fungal dataset introduced in Section 2.1, tested across different architectures. The approach effectively scales across architectures with few training epochs.

\textbf{Model Performance}:
The proposed model was fine-tuned using a batch size of 512 on an NVIDIA GeForce RTX 4090. The fine-tuned CLIP model was trained on a dataset comprising fungal images and LLM-generated textual descriptions, aiming to distinguish key biological structures. The synthetic fungal dataset used in this study consisted of 6,000 images, split into 4,800 for training, 600 for testing, and 600 for validation, and was trained for 100 epochs. The fine-tuning (FT) strategy of pairing image-text samples within classes provides a streamlined and efficient approach for zero-shot fungal classification.

Initially, CLIP’s parameters were frozen to preserve its broad representations from large-scale pretraining. Transformer layers were incrementally unfrozen every five epochs, enabling a progressive FT process. This stepwise unfreezing allowed the model to adapt to the synthetic fungal dataset while maintaining stable learning. Layers were unfrozen in reverse order, starting from the last transformer block, ensuring that lower-level transferable features remained intact. This approach led to better convergence while optimizing computational efficiency, making the FT process feasible even with limited resources. The AdamW optimizer was applied exclusively to the unfrozen layers, ensuring efficient parameter updates.

Inference on the validation set produced promising classification results, with the model assigning high-confidence predictions to the correct labels in most cases. The probability distribution of predictions revealed that the model effectively differentiated between hyphae, spores, and mycelium with reasonable certainty. However, some mis-classifications were observed, likely due to overlapping visual features among specific categories. Figures~\ref{Fig4} illustrate the zero-shot classification performance of the fine-tuned CLIP model when paired with textual descriptions from different LLMs: GPT-4o1, Claude 3.5, and Gemini 2.0. The CLIP model demonstrated strong recognition capabilities for fungal growth stages across various LLM-generated descriptions. High similarity scores between text and image embeddings indicate that these prompts effectively capture fungal structural characteristics. However, descriptions from Claude 3.5, while informative, were less aligned with the synthetic fungal dataset. Furthermore, Gemini 2.0 exhibited the lowest classification accuracy, suggesting its textual outputs lacked the specificity required for optimal alignment.

These findings emphasize the importance of well-structured, domain-specific textual descriptions in bridging the modality gap. The results further indicate that text quality significantly influences classification accuracy, with GPT-generated captions yielding the most precise alignments. This suggests that incorporating high-quality, domain-specific textual representations can enhance fungi classification performance.

\begin{figure}[H]
\centering
\includegraphics[width=\linewidth]{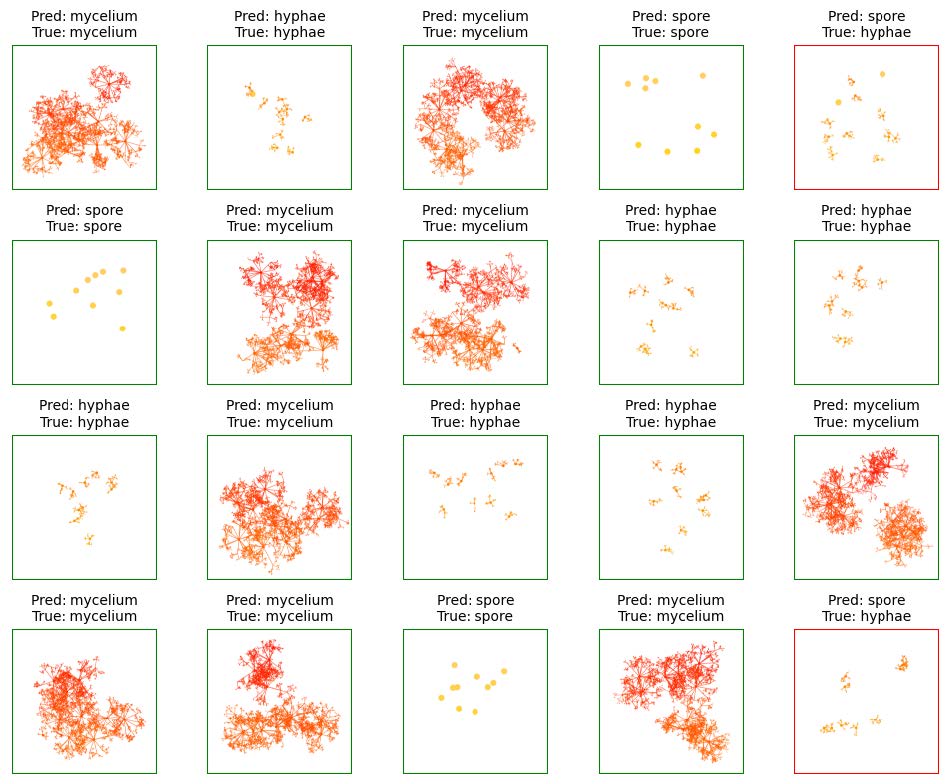}
\caption{Recall@1 qualitative results. The predicted and true class labels from the validation dataset have been mentioned in this figure. The results which are correctly ranked as 1 for their corresponding class label are bordered green while incorrect results have been bordered red.}
\label{Fig5}
\end{figure}

\textbf{Performance Across Architectures}:
Table \ref{tbl1} presents the classification performance of different CLIP model architectures trained on the synthetic fungal dataset:
\begin{itemize}
\item ViT-L/14@336px: These architectures achieved the highest Recall@1 score of 0.97, indicating high performance in fungal image-text alignment. The results suggest that increasing model complexity leads to significant accuracy gains for the fungi dataset.
\item ViT-B/32 and ViT-B/16: This model demonstrated a Recall@1 score of 0.67 and 0.91, respectively. 
\item RN50: This model exhibited the lowest Recall@1 score (0.333), indicating that convolution-based architectures may be less effective in capturing fine-grained fungal features.
\end{itemize}
These results confirm that transformer-based architectures, particularly ViT-based CLIP variants, are better suited for zero-shot fungal classification tasks.

\textbf{Error Analysis}: Figure~\ref{Fig5} visualizes the predicted class labels from the validation dataset, providing insight into the model’s classification behavior. The model exhibited high confidence in correctly classifying spores and hyphae. However, misclassifications were observed between hyphae and mycelium due to their overlapping structural characteristics. Lower confidence scores were recorded for ambiguous samples, highlighting areas where text-image alignment could be further refined, as discussed in Figures~\ref{Fig4}.

Overall, the results indicate that ViT-L/14@336px achieved the best classification accuracy, demonstrating strong alignment with the synthetic fungal dataset. The effectiveness of textual descriptions was evident, with GPT-4o1 outperforming Claude 3.5 and Gemini 2.0 in generating highly relevant prompts that enhanced classification performance. Misclassifications primarily occurred in visually similar growth stages, suggesting that refining text-image alignment could further improve accuracy. These findings highlight the potential of integrating synthetic datasets with LLM-generated text to advance zero-shot fungal classification.

\section{Conclusion and Future Work}
This study demonstrates the effectiveness of enhancing zero-shot classification in vision-language models (VLMs) like CLIP by integrating synthetic fungal images and LLM-generated textual descriptions. By aligning these complementary datasets within CLIP’s shared representation space, we successfully improved classification performance across different fungal growth stages. Fine-tuning CLIP with domain-specific synthetic data enhanced generalization and minimized modality gaps, while the inclusion of diverse LLM-generated text significantly refined classification accuracy. Additionally, transformer-based architectures, particularly ViT-L/14@336px, exhibited superior performance compared to convolution-based models like RN50, confirming their suitability for fine-grained fungal classification.

Despite these advancements, several areas remain for future improvement. Expanding the dataset to include additional fungal growth stages and diverse environmental conditions could improve model generalization. Enhanced text representations, such as richer text embeddings or multi-label classification, may help mitigate ambiguities in class descriptions, particularly in visually similar categories like hyphae and mycelium. Moreover, FT with domain-specific loss functions, such as contrastive learning tailored for biological classification, could further strengthen the semantic alignment between image and text embeddings. Another promising direction is refining the model’s robustness against classification errors by improving text-image alignment strategies and leveraging more structured textual prompts. Additionally, the progressive fine-tuning approach used in this study, where transformer layers were gradually unfrozen, proved effective in optimizing computational efficiency while maintaining stable learning. Future work could explore adaptive fine-tuning strategies that dynamically adjust layer unfreezing based on validation performance, further reducing computational overhead while enhancing model adaptation.

Overall, this research contributes to advancing automated fungal identification, with potential applications in micro-biological monitoring studies. Future refinements in dataset diversity, hierarchical label structures, and optimization strategies could further strengthen zero-shot classification capabilities in specialized biological domains.

%\printcredits
\section*{Ethics Statement}
This research does not involve experiments, observations, or data collection related to human or animal subjects. 

\section*{Declaration of competing interest}
The authors declare that they have no known competing financial interests or personal relationships that could have appeared to influence the work reported in this paper.

\section*{Data Availability}
\href{https://data.mendeley.com/datasets/rw6ndgyrd7/1} {Synthetic Fungi Generation}. 

\bibliographystyle{unsrtnat}
\bibliography{references}  
\end{document}